# A Comparison of Meta-heuristic Search for Interactive Software Design

C.L. Simons and J.E. Smith

Department of Computer Science and Creative Technologies,

University of the West of England, Bristol, BS16 1QY, United Kingdom.

+44.117.3283135, +44.117.3283161

{chris.simons, james.smith}@uwe.ac.uk

**Abstract** Advances in processing capacity, coupled with the desire to tackle problems where a human subjective judgment plays an important role in determining the value of a proposed solution, has led to a dramatic rise in the number of applications of Interactive Artificial Intelligence. Of particular note is the coupling of meta-heuristic search engines with user-provided evaluation and rating of solutions, usually in the form of Interactive Evolutionary Algorithms (IEAs). These have a well-documented history of successes, but arguably the preponderance of IEAs stems from this history, rather than as a conscious design choice of meta-heuristic based on the characteristics of the problem at hand. This paper sets out to examine the basis for that assumption, taking as a case study the domain of interactive software design. We consider a range of factors that should affect the design choice including ease of use, scalability, and of course, performance, i.e. that ability to generate good solutions within the limited number of evaluations available in interactive work before humans lose focus. We then evaluate three methods, namely greedy local search, an evolutionary algorithm and ant colony optimization, with a variety of representations for candidate solutions. Results show that after suitable parameter tuning, ant colony optimization is highly effective within interactive search and out-performs evolutionary algorithms with respect to increasing numbers of attributes and methods in the software design problem. However, when larger numbers of classes are present in the software design, an evolutionary algorithm using a naïve grouping integer-based representation appears more scalable.

## 1 Introduction

The application of automated search to a range of software development activities has attracted significant research attention. Indeed, Search-Based Software

Engineering (SBSE) (Harman 2007, Harman 2011) is now a well-established discipline. SBSE historically focused on software testing where solutions can be represented fairly naturally and metrics such as structural and functional test coverage can be automatically calculated to serve as quality functions. However, in the upstream stages of the software design, such as the object-oriented modeling of design classes, the choice of evaluation functions is much less well defined – for example, Bowman *et al.* (2010) cite 6 different possible metrics relating to the structural integrity of the design with respect to design coupling and cohesion. Here the precise balance of factors affecting the subjective judgments of the human software engineer is less well understood – hence the oft-heard references to the "art" of software design. Indeed, this is precisely the sort of scenario in which Interactive Evolutionary Algorithms (IEAs) have been shown to perform well (see e.g. the survey in Takagi (2001), and more recent work such as Tagaki and Ohsaki (2007), Celeb-Solly and Smith (2007), Brintrup et al. (2008) and Pauplin et al. (2010) ). Our earlier work demonstrates that we can indeed successfully use meta-heuristics to provide computational support for an interactive software design process, evolving object-oriented class models that met designers' criteria –both subjective (Simons et al. 2010) and aesthetic (Simons and Parmee (2012).

As with most papers in the field, such interactive design search uses an Evolutionary Algorithm (EA) (Eiben and Smith, 2003) because of their long history of successful applications. However, as the name of the field of Search-Based Software Engineering suggests, potentially any search algorithm could be used, although in practice research effort has also tended to concentrate on meta-heuristics, in particular Evolutionary Algorithms. It is appropriate that we challenge adoption of a technology based on history, and examine whether other search methods might be better suited to some, if not all, interactive design search tasks. Indeed the same argument has been made for SBSE in general: "*We must be wary of the unquestioning adoption of evolutionary algorithms merely because they are popular and widely applicable or because, historically, other researchers have adopted them for SBSE problems; none of these are scientific motivations for adoption.*" (Harman 2011).

One major contribution of this paper is to identify a number of factors that we believe are crucial to making an informed choice for an underlying search engine for interactive search (Section 4). Then, to make the comparison concrete, we describe the experimental methodology followed and three case studies of early lifecycle software design tasks (Section 5). Results of comparing the different algorithms according to the factors identified are presented in section 6, and finally in section 7, we conclude by making some recommendations for possible users of interactive search tools.

## 2 Background

### 2.1 Search-Based Software Engineering

From its early roots using genetic algorithms to evolve software test sequences (e.g. Xanthakis et al. 1992, Smith and Fogerty 1996, Jones et al. 1996) and microprocessor design verification tests (Smith et al. 1997), the idea that parts of the software development process are essentially optimization problems, and as such are amenable to automated search, has rapidly gained currency. In most cases the search suffers from combinatorial explosion, and the "fitness" landscapes are thought to exhibit discontinuities and multiple optima, motivating the use of meta-heuristics such as evolutionary algorithms to perform the search. The term "Search Based Software Engineering" (SBSE) was coined around the turn of the millennium by Harman and Jones (2001). In the last decade applications of SBSE can be found across the spectrum of the software engineering lifecycle, including requirements analysis and scheduling (Ren et al. 2011), design tools and techniques (Bowman et al. 2010, Simons et al. 2010), software testing (McMinn 2004) and automated bug fixing (Weimer et al. 2010), and software maintenance (O'Keefe and Cinneide 2008). Harman (2011) overviews how software engineering and evolutionary computing have come together, while a comprehensive repository of publications in SBSE is maintained by Zhang (2012).

### 2.2 Object-Oriented Software Design

The first stage in software design is to identify and evaluate the concepts and information relevant to the design problem domain under investigation. Using the object-oriented paradigm, such concepts and information from the design problem domain are expressed using the 'class' construct, where individual instances of classes are known as objects. These classes and objects have crucial relevance to subsequent downstream software implementation and testing. The Unified Modeling Language (UML) (Object Management Group, 2012) is the standard modeling language of the object-oriented paradigm, and is widely used by software designers to visualize and specify classes as well as other aspects of software designs. Using the UML, classes are placeholders or groupings of attributes (i.e. data that need to be stored, computed and accessed), and methods (i.e. units of execution by which objects communicate with other objects or indeed with human users, other programs etc.) Thus early lifecycle software design involves finding an appropriate grouping of attributes and methods into classes. We will henceforth refer collectively to methods and attributes as *"elements"*, and candidate groupings of elements into classes as *"designs"*. To ensure that each class grouping is meaningful as a relevant concept to human designers, we impose the constraint that each class grouping contains at least one attribute and at least one method. Formulating software design as a search problem, an instance of the class modeling problem can be represented in a number of ways. Common to all these is that to enable efficient search it is necessary to assign a quality measure to candidate design solutions. A large number of quantitative metrics have been identified in the software design community, many referring in different ways to the structural integrity of the design with respect to "coupling" (the extent to which one class depends on others to fulfill its capabilities) or "cohesion" (the extent to which a class has clear purpose) e.g. Harrison et al. 1998, Briand et al. 1999, Al Dallal and Briand 2010. It is generally held that a good software design should exhibit low coupling and cohesion, but these are potentially competing measures. Indeed, this is a typical example of how many design decisions can be balancing acts, trading-off one quality measure against another. This has led some authors to use multi-objective, quantitative approaches to search e.g. Bowman et al. 2010, Harman and Tratt 2007. However, our work has taken a different approach: rather than attempt to define coupling/cohesion metrics which capture what a user is looking for, and then manage the quantitative multi-objective trade-off, we have used a multi-objective IEA where the designer is responsible for assigning qualitative fitness to a candidate solution (Simons et al 2010). To relive the burden of interaction fatigue on users, we have also investigated the use of a surrogate fitness function that attempts to learn a model of qualitative "elegance" from the users' decisions (Simons 2011, Simons and Parmee 2012).

A number of approaches present as possibilities for application to object-oriented software design search. For example, EAs have been used with some success for general "grouping" problems (Falkenauer 1998, Tucker et al. 2006), but there is still considerable debate over the best choice of representation to avoid massive redundancy in the phenotype / genotype mapping, and this remains a major unsolved problem (Lewis and Pullin 2011). Inherently the problem stems from the fact that, for example, if two elements *i* and *j* should be co-located within a class, then not only is the choice of label for that class irrelevant, but also adapting an EA to account for this representational constraint is non-trivial and at best creates a highly specialized algorithm. Given that the evolving population represents a probability distribution function of the assignment of elements to classes, a further possibility might be to use an Estimation of Distribution Algorithm (EDA) (Lozano et al. 2006). For a graphical model that correctly captured the grouping above, then trivially the probabilistic model could evolve to look like $P(i = k \mid j = l) = \delta_{kl}$, where $\delta_{ij} = 1$ if $i = j$ and 0 elsewhere. However, currently EDAs, like other probabilistic model builders, use greedy search to construct models, so the search will at best be as effective as a greedy local search algorithm in the space of partitions. Furthermore, in contrast, Ant Colony Algorithms (ACOs) (Dorigo and Stutzle 2004) have been used very successfully for problems with an inherent grouping component such as the Vehicle Routing Problem (VRP) (Toth and Vigo 2011) since the pheromone trail (broadly equivalent to the probabilistic graphical model in an EDA) can effectively contain a set of partial paths to be selected and traversed by ants without need for class labels, hence avoiding the whole issue of redundancy.

Given the existence of many contrasting search approaches to the object-oriented software design problem, one factor which motivates their evaluation is the need to manage a range of instances of realistic scalability. Especially in interactive mode, there are clear benefits from being able to ability to rapidly visualize and "freeze" certain components. This is far more easily achieved in some meta-heuristics (e.g. by setting the pheromone trails to artificially high values) than others (e.g. modifying variation operators in an EDA to preserve portions of a genome).

## 2.3 Interactive Meta-Heuristic Search

Interactive EAs were popularized in Dawkins' 'biomorphs' program (1990), but build on a well-established field in Artificial Intelligence. They have been successfully applied in a wide range of applications to facilitate user-personalization without the need for time consuming explicit knowledge-acquisition process (Tagaki 2001). Typically the user is presented with a number of solutions, and rates them according to the extent to which they match the user's desiderata. Thus this process implicitly captures the user's multi-objective decision making processes. Well known early applications include face-recognition (Caldwell and Johnson1991), the evolution of computer graphics (Sims 1991a, 1991b), fitting Cochlear Implants (Legrand et al. 2007) and hearing aids (Ohsaki et al. 1998).

A common feature of IEAs is their reliance on human guidance and judgment to direct and control the search which creates both potential weaknesses and strengths. On one hand, human assessment tends to have a component of subjectivity and non-linearity of focus over time. Thus including a human in-the-loop introduces a need for rapid convergence to prevent the interactive process from becoming tedious for the human participant. At the same time the ability to manoeuvre the search interactively can potentially be exploited as a powerful strategy for adapting an otherwise naive EA.

There have been a number of studies addressing the issues related to minimizing fatigue both, physical and psychological, that can result from prolonged interaction times and the possible stress of the evaluation process. Discretizing continuous values to using five or seven levels was shown to facilitate decision making when allocating fitness values, without the quantization noise significantly compromising convergence (Ohsaki et al. 1998), and this limit on capacity for processing information has been comprehensively discussed in Millar (1956) where he suggests organizing the information into several dimensions and successively into a sequence of "chunks" could help stretch this limit on bandwidth.

Alternative ways of reducing time taken to discover good solutions is by considering larger population sizes and using a screening mechanism whereby only a few individuals showing good fitness are displayed to the user. Several methods have been proposed to approximate fitness by, for example, clustering individuals (Lee and Cho 1999, Boudjeloud and Poulet 2005) or using multiple fuzzy state-value functions to approximate the trajectory of human scoring (Kubota et al. 2005). An interactive concept-based search using a multi-objective evolutionary algorithm was proposed by Avigad at al. (2005) which combined a model-based fitness of sub-concept solutions (using a sorting and ranking procedure) with human evaluation. The efficacy of combining qualitative (user-provided) and quantitative (computer-generated) objectives was also demonstrated in Brintrup et al. (2008). Within SBSE, design elegance has been exploited as a model and surrogate fitness function that then is dynamically combined with quantitative objectives to produce elegant software designs (Simons and Parmee, 2012).

Surprisingly, the literature does not seem to contain many examples of the user of alternatives to EAs as the underlying heuristic for interactive search. Rather, approaches rely on using either a meta-heuristic with a defined quality function and periodically using user interaction to guide search by reformulating a fitness function or preference weighting, or simply to change the search characteristics via changes to the algorithm parameters. Examples of the former include multi-objective Iterated Local Search (Geiger 2008) and Tabu Search (Kopfer and Schonberger 2002) and of the latter include Ant Colony Optimization (Ugur and Aydin 2009). Interestingly, however, there is one report of interactive search with Particle Swarm Optimization used to design temperature profiles for a batch beer fermenter (Madar et al. 2005).

## 3 Meta-heuristic Search Algorithms

### 3.1 Representations

In the first representation, which we shall call "naïve grouping" (NG), the genotype is a sequence of $d$ integers from the set $\{1,\ldots,c\}$, (where $c$ is the maximum number of classes allowed), with an allele value of grouping $g_i = j$ being interpreted as putting element $i$ into class $j$. The search space is of size $c^d$, but there is considerable redundancy in the genotype-phenotype mapping since as far as the class model is concerned; the label applied to a class is irrelevant.

The second representation is an Extended Permutation (XP) inspired by the Travelling Salesman Problem (TSP) and Vehicle Routing Problem (VRP) (Toth and Vigo 2001). Candidate solutions are represented as permutations of a set of ($e+c$) elements, where $e$ are the attribute and method elements and the extra $c$ elements are interpreted as "end of class" markers akin to a "return to depot" in a VRP instance.

## 3.2 Fitness Measures

To reflect the interactive nature of the meta-heuristic search, a combination of fitness measures is used. It is generally understood in software engineering that to achieve structural integrity designers strive for high cohesion in classes (to reflect a clear purpose) and low coupling between objects (to ensure the design is robust yet flexible to change). Therefore, firstly, the measure of the structural integrity chosen is inspired by the "Coupling Between Objects" (CBO) measure (Harrison et al. 1998). Regardless of the representation chosen, each candidate solution is decoded into a set of classes, and the CBO is calculated as the proportion of all uses of attributes by methods that occur across class boundaries. This is expressed as a maximization function $f_{CBO}$ = (1.0 -CBO) x 100, so that $f_{CBO}$ = 100.0 for a completely de-coupled design (all uses occur inside classes) and 0.0 for a completely coupled design.

However, it is also necessary to reflect the interaction of the designer within search. We have previously found that the elegance of the software design has proven to be a useful interactive measure (Simons and Parmee 2012), and proposed a number of quantitative elegance metrics relating to the evenness of distribution of attributes and methods among classes within the design. Building on this, two elegance metrics have been chosen as surrogates for human qualitative elegance evaluation, namely:

- *Numbers Among Classes (NAC)*: the standard deviation of the numbers of attributes and methods among the classes of a design. This was truncated to the range [0,R] and a fitness to be maximised calculated as denoted $f_{NAC}$ = 100* (R- NAC)/R. The higher this value, the more symmetrical the appearance of attributes and methods among the classes in the design.

- *Attribute to Method Ratio (ATMR)*: the standard deviation of the ratio of attributes to methods for each of the classes in a design. A fitness $f_{ATMR}$ was calculated in the same way as above. The higher this value, the more even and symmetrical the appearance of this ratio across individual classes of the software design.

### 3.3 Evolutionary Algorithm

The EA chosen for comparison uses deterministic binary tournaments for parent selection and a generational replacement model ensures the search is comparable to ACO. Random uniform mutation with either One-Point or Uniform crossover is applied to the NG representation. For the XP representation, we used Order-based crossover (Davis 1991) and "Edge Recombination" (Mathias and Whitley 1992). The former preserves the relative order of elements (as per scheduling type problems) and the latter preserves adjacency information (as per TSP or VRP). We have previously shown that for many permutation-based problems the choice of mutation operator depends on both the problem instance and the state of the search (e.g. Krasnogor and Smith 2001, Serpell and Smith 2010 for adjacency-, and Smith et al. 2009 for order-based problems). Therefore fixed mutation rates for the EA are interpreted as either the locus-wise probability of randomly resetting an allele value (NG) or as the probability of applying a single mutation event of type 'Swap', 'Insert' or 'Invert' chosen at random (XP). In both cases we also examined the utility of self-adaptation to provide robust optimization performance, and reduce the number of parameters required. Following the schemes in Smith 2001, Stone and Smith 2002, Serpell and Smith 2010, a single extra gene is used to encode for one of a set of possible mutation rates. During mutation, first the encoded value is randomly reset with probability 0.1, then a mutation event occurs in each locus with the encoded probability.

### 3.4 Ant Colony Optimization

ACOs have been used successfully for permutation-type problems (Toth and Vigo 2001, Dorigo et al. 2006) where the pheromone trails map naturally onto path-based problems such as the TSP and VRP. Therefore it was natural to use the XP representation described above. The ACO has been implemented as described in Dorigo and Stutzle (2004). Each ant creates a solution by visiting elements (attributes, methods or "end of class") in turn, choosing each element probabilistically according to a combination of the attractiveness ($\alpha$) of pheromone trails (laid down by previous ants) and heuristics. After the whole population (colony) has created tours, all pheromone trails are subject to evaporation at a constant rate ($\sigma$). Finally, for each link traversed in each of the trails, a small amount ($\mu$) of additional pheromone is laid down proportional to the fitness of the trail in which it occurred.

### 3.5 Greedy Local Search

In keeping with the focus on re-using of-the-shelf components, Greedy Local Search (GLS) is implemented as (1+1) version of the EA code with mutation replaced by systematic (rather than randomized) search of the NG values or the 2-opt neighborhood for the XP representation

## 4 Factors Affecting Choice of Meta-heuristic Search Engine

We identify a number of factors that we believe should be considered when choosing the meta-heuristic as part of creating an interactive tool. Some of these factors lead themselves to be easily quantified, other less so. Without wishing to second guess the uses to which interactive optimization can be turned, we do not attempt to rank these according to importance. Rather, we present them using examples of software design problems to illustrate the issues involved.

### 4.1 Scalability

When we consider scalability, we mean not just how well do the algorithms scale to solve larger problems, but how well can they assist the human in the process of designing large software solutions. Typically this process will involve partitioning the problem in some way to reduce its dimensionality. Key factors therefore include how well the algorithms support the user in first identifying good partial solutions, and then "freezing" those partial solutions. Clearly this depends on the complexity of the mapping from candidate solution (as presented to the user for evaluation) back to representation (as used by the search engine). For a software design, it is fairly simple to imagine an interactive box whereby a user can select a class to "freeze". For an EA, or Local Search based method, this requires some method for permanently recording the information that certain genes should not be affected by mutation/perturbation and should be co-transmitted under recombination. However some meta-heuristics explicitly perform this partitioning process and so intrinsically record this information. Thus this "freezing" process can be instantiated by simply "fixing" some elements of the graphical model in an EDA or setting the relevant pheromone levels to an arbitrary high value in an ACO.

The other aspect of scalability for interactive meta-heuristic search is of course how well a given algorithm scales without the freezing process. Given the limits on human attention, this relates to the ability to discover high quality solutions in relatively few evaluations as the dimensionality of the problem increases. It is probably pointless to attempt to draw any firm conclusions about the relative scalability of different methods, as this is of course entirely problem and parameter dependent, although recent theoretical results (Birattari et al. 2007) suggest how to avoid poor scalability in ACO which was previously thought to be a problem. However the need to make rapid advances in fitness rather than evaluating randomly created solutions points to either the user of local search, or to small populations in EAs or ACOs. EAs are known to work well with small populations (e.g. CMA-ES for continuous optimization), this is less well examined for ACOs. One major factor that should be considered is the number of human interactions required, and the availability (or otherwise) of surrogate fitness function that can reduce this.

**4.2 Robustness**

Meta-heuristic robustness relates to a number of factors e.g.

- *Appropriateness of representation:* how sensitive and appropriate is the representation? For example, might a permutation representation cause problems of degeneracy?

- *Support for search:* how well do the algorithms support for single- and multi-objective search?

- *Parameter choice:* is algorithm performance effective across a range of parameters?

- *Parameter tuning / self-adaptation:* can parameters be automatically tuned and/or controlled?

## 4.3 'Off-the-shelf' Availability

A number of frameworks and toolkits for implementing EAs are readily and freely available, in all of the major programming languages and environments. Most of these can be adapted to run with a parent population of size one in order to implement local search. Well known toolkits such as Evolving Objects (Keijzer et al. 2002) and ECJ (Luke et al. 2012) implement a range of different data types and provide sufficient different mutation and recombination operators to give the user considerable flexibility in their choice of problem representation. Similarly, a number of implementations of the ACO meta-heuristics for optimization (rather than data mining) are available in C and C++ programming languages via the ACO website (2012). Although a range of different algorithm variants are supported and versions for different problems are available, all assume the use of a permutation representation for solutions. This restriction of the available ACO implementations, and the fact that most papers in this field deal with a path-type representation, would appear to rule out the straightforward use of ACOs for some problems. After consideration, all of the meta-heuristic approaches considered in this paper were implemented by the authors taking specifications from the literature. Assuming a reasonable knowledge of software engineering, the available implementations could be adapted to other problems (for example the heuristic rules used to initialize and augment the pheromone trails) but the level of documentation and support is not as comprehensive as might be expected from the relative maturity and popularity of the field – which should not be read to reflect the scientific merit.

## 4.4 Constraint Handling

A crucial factor in the choice of meta-heuristic is the ease (or difficulty) with which the algorithm can handle any domain-specific constraints. For the software design problem, there is one crucial constraint as described earlier: each class must contain at least one method and one attribute. This constraint has a significant impact on the grouping of elements (attributes and methods) to classes, and the ease (or difficulty) with which the meta-heuristic copes with this is a key factor in its choice.

# 5 Methodology

## 5.2 Strategy

A key factor in the comparison of the meta-heuristic algorithms is the availability of plausible and representative test design problems for early lifecycle software design. Unfortunately, benchmark software design problems do not appear readily in either the research literature or industrial repositories. Therefore, three real-life software design problems have been selected for use. These have been chosen to provide an appropriate range of problem domain, and scale. While it is not possible to precisely assess how representative these might be of the software design field as a whole, both the second and third problems have been drawn from fully enterprise scale industrial software developments, and are decidedly non-trivial in size and complexity. Details of the three design problems are given in the following section. For the benefit of the community, full problem specifications, source code for all the algorithms used, and all results are made available on line (Simons and Smith, 2012).

To permit large scale comparisons we took a two-stage approach. Firstly, to establish the sensitivity of metaheuristic's performance to parameter values, the number of constraints, and how they are handled, we focussed on Coupling Between Objects, optimizing $f_{CBO}$ and comparing with manually produced designs. In this stage we used all combinations of the parameter values in Table 1.

Secondly, using the "best" parameter sets established for each method, we simulated multi-objective interactive search by introducing the surrogate elegance metrics into a weighted-sum approach and optimising: $f_{MO} = a. f_{CBO} + b. f_{NAC} + c. f_{ATMR}$. Empirical calibration revealed that a weight of $a = 0.8$ was required for CBO, emphasising the importance of design structural. To reflect the inherently noisy nature of human evaluation, we then chose *b* uniformly from the interval (0, 1-*a*) and set *c = 1.0 – a - b*.

All runs use a fixed number of classes – the same as in the manual design solution to provide comparability. To ensure repeatability of results, we made 50 runs for each test, i.e., each combination of algorithm, problem, encoding, and parameter values. Each run is allowed to continue until either one million solutions were evaluated, or a software design with fitness 100.0 was discovered. For each run we recorded the values of $f_{CBO}$, $f_{NAC}$, and $f_{ATMR}$ for best solution found and the number of solutions evaluated before this best solution is first discovered. These are denoted MBF and AES respectively.

Table 1: Search Parameters for Meta-heuristic Algorithms

| | Parameter | Values trialled |
|---|---|---|
| Stochastic Local Search | Perturbation method | NG: Allele-wise mutation. XP: random one of insert/invert/swap mutation |
| Evolutionary Algorithm | Selection Method | Tournaments to select parents, size 2 and 5. Generational replacement with elitism. |
| | Crossover Probability | 0.0, 0.2, 0.4, 0.6, 0.8, 1.0 |
| | Crossover Operator | NG: Uniform, One Point XP: Order-based, Edge Recombination |
| | Mutation Probability | Self-adaptive: 1/*l* *(0.001, 0.002, 0.01, 0.02, 0.01, 0.2, 1, 2, 5, 10, ) Fixed: 0.001, 0.01, 0.05, 0.1, 0.25, 0.5 |
| | Population Size | 25, 100 |
| Ant Colony Optimisation | Trail Attractiveness ($\alpha$) | 0.0, 0.5, 1.0, 1.5, 2.0, 2.5, 3.0 |
| | Pheromone Update ($\mu$) | 0.0, 0.5, 1.0, 1.5, 2.0, 2.5, 3.0, 3.5 |
| | Pheromone Decay ($\rho$) | 0.0, 0.01, 0.1, 0.25, 0.5, 1.0 |
| | Ant Colony Size | 25, 100 |

Wherever results are analyzed by comparison of means, the "General Linear Model" of IBM SPSS Statistics tool version 19 is used with algorithm choice, population size and design problems as fixed factors, then applying ANOVA followed by post-hoc testing using Tukey's "Honestly significantly different" test. In what follows, statements that effects are significant or not, should be read to mean that they are statistically significant with over 95% confidence according to these tests.

## 5.2 Software Design Problems

Three software design problem domains are used as vehicles for investigation. The first is a generalized abstraction of a Cinema Booking System (CBS), which addresses, for example, making an advance booking for a showing of a film at a cinema, and payment for tickets on attending the cinema auditorium. A specification of the use cases of Cinema Booking System design problem is available at Simons (2012a). This problem has 16 attributes, 15 methods and 39 method/attribute uses. The second software design problem domain is an extension to a student administration system performed by the in-house information systems department at the University of the West of England, UK. The university sought to record and manage outcomes relating to the Graduate Development Program (GDP) of students during their studies. The extension was implemented and deployed in 2008. A specification of the use cases used in the development is available from Simons (2012b). The GDP problem has 43 attributes, 12 methods and 121 method/attribute uses. The third software design problem domain is based on an industrial case study – Select Cruises (SC) - relating to a cruise company selling nautical adventure holidays on tall ships where passengers are members of the crew. The resulting computerized system handles quotation requests, cruise reservations, payment and confirmation via paper letter mailing. A specification of the use cases of Select Cruises design problem is available at Simons (2012c) The SC problem has 52 attributes, 30 methods and 126 method/attribute uses.

Manual designs created by the appropriate software engineers for the three problems are available at Simons (2012d). Numbers of classes and $f_{CBO}$, $f_{NAC}$ and $f_{ATMR}$ (R=6.0) for the manual designs are given in Table 2.

Table 2: Values of measures of Manual Software Designs

|     | Number Of Classes | Coupling Between Objects | $f_{CBO}$ | Elegance $f_{NAC}$ | Elegance $f_{ATMR}$ |
|-----|---|-------|------|-------|-------|
| CBS | 5 | 0.154 | 84.6 | 86.31 | 96.69 |
| GDP | 5 | 0.297 | 70.3 | 56.80 | 56.38 |
| SC  | 16 | 0.452 | 54.8 | 74.67 | 69.2 |

# 6 Single Objective Results

We begin this section with an assessment of the sensitivity of the two population based methods to their parameters. In general constraints can be handled directly (i.e. via repair mechanisms, specialized operators, or decoders) or indirectly via penalty functions (Eiben and Smith 2003). The former can be more efficient, but require problem specific alteration of the algorithms [ibid]. Therefore initially constraints were handled indirectly, by setting the fitness to zero for all solutions containing classes that had zero methods or zero attributes. This is followed by an analysis of a relatively minor change to each method so that new candidate solutions were regenerated (by following pheromone trials or via recombination and mutation) if they contained invalid classes – effectively a very crude direct method. We end by directly comparing the results from the "best" parameter sets of three meta-heuristics. For the sake of clarity we have summarized the results obtained, but full details, statistical analysis and result logs may be found with the source code and problem definitions at Simons and Smith 2012.

## 6.1 Greedy Local Search

Table 3: Mean Best Coupling and Number of Evaluations to Best Solution for Greedy Local Search. Results shown for all, and for just successful runs. Standard Deviation is shown in parentheses.

|     | Manual $f_{CBO}$ | Encoding | MB $f_{CBO}$ -all | N valid | MB $f_{CBO}$ -valid | AES |
|-----|---|---|---|---|---|---|
| CBS | 84.6 | NG | 87.25 (3.16) | 50 | 87.25 (3.16) | 62669 |
|     |      | XP | 62.35 (35.7) | 38 | 82.04 (5.35) | 5272 |
| GDP | 70.3 | NG | 87.35 (4.05) | 50 | 77.35 (4.05) | 76088 |
|     |      | XP | 57.61 (36.46) | 37 | 77.85 (13.80) | 13428 |
| SC  | 54.8 | NG | 60.89 (27.1) | 42 | 72.49 (4.07) | 598493 |
|     |      | XP | 0.0 (0.0) | 0 | - | - |

Table 3 shows the results obtained with GLS under the two encodings. For all problems with XP, and for the SC problem with NG representation, some runs failed to find valid solutions, and the second column of fitnesses, and the AES column reflect only those runs funding valid solutions. Whether considering all, or only successful runs, the quality of solutions found by GLS algorithm is significantly higher on the NG landscape than on the XP one. There is also a far higher variance in the quality of local optima found on the XP landscape. GLS finds solutions with lower coupling than the manual design – but this perhaps merely emphasis the multi-objective nature of human design. It is worth commenting that the comparative values of AES on the two landscapes - path from a random starting point to the local optimum are typically an order of magnitude longer on NG than XP landscape, although of course this does not necessarily mean that good quality solutions are not discovered early on the way. We can understand this by noting that both representations contain a certain amount of redundancy. On the NG landscape there are $c!$ different labellings of any given arrangement of elements into c classes. However since the order of the tours is immaterial, the XP representation is far more redundant since if we denote the size of each class as $|c_i|$ (I = 1,…,c) then each design can be represented by $\prod_i |c_i|$ paths, so there are far more local optima in the XP landscape. Looking at the progress of sample runs, the number of solutions evaluated to reach given for XP and NG are 142/182 (fitness 50, CBS) and 480/839 (fitness 75 CBS), 491/491 (fitness 50, GDP) and 861/853 (fitness 75, GDP). On the SC problem the sample run of GLS-NG evaluated 372006 solutions before finding a valid one and then a further 2674 before finding one with fitness 50, i.e., comparable to the manual design .

Overall, the high failure rates and variability of end results suggests that GLS-XP is unsuited to interactive search. The failure of some GLS-NG runs on SC shows that a specialized initialization operator is needed, and the subsequent long time to find a human-comparable design on SC, suggests that even if one is available, the GLS algorithm may be unsuitable for interactive search.

### 6.2 Evolutionary Algorithms

Analysis of the results from EAs with both representations confirmed that the fixed mutation rate which gave the highest $f_{CBO}$ values depended on both the problem (hence representation length) and on number of factors which affect the exploration-exploitation balance (population size, crossover operator and probability), and sub-optimal choices greatly deteriorated performance. However, on a more positive note, results also clearly showed that for every problem-representation pairing, the use of Self-Adaptation leads to the discovery of solutions with significantly higher fitness than any of the fixed mutation rates, without any significant penalty in terms of the number of evaluations taken. Furthermore, when using self-adaptive mutation neither the choice of tournament size nor of crossover probability (within the range 0.2-0.8) made any significant difference to the $f_{CBO}$ values. Since self-adaption not only yields superior results, but also increases the EAs robustness by reducing the number parameters required, for brevity we only report these results henceforth.

Figure 1 shows mean best coupling achieved with self-adapting mutation, 60% probability of applying crossover for four crossovers and two population sizes. As can be seen, for the CBD and GDP problems there is little difference between population sizes, although the larger is beneficial for SC, the gains are never statistically significant with NG representation. With XP the effect of population size on the SC problem is significant, and relates to the initialization problem as with GLS.

From the perspective of the robustness of the algorithms, there is a positive outcome in that the same settings do well across all three problems and on almost every run the EAs discover solutions with higher $f_{CBO}$ than the equivalent human crafted solution. Only on the largest scale SC problem did EAs using the permutation representation fail to beat the human-crafted values.

Immediately apparent, and confirmed by statistical analysis, is that the NG (One point and Uniform recombination) representation leads to the discovery of better solutions than XP ( Edge or Order recombination). For the SC problem, the difference is typically 50%. Factoring out the effect of problem instance, there is not a significant difference between Uniform and One Point recombination for the NG representation. However, with the permutation-based XP, use of the Order crossover discovers higher quality solutions than Edge Recombination. Note that the latter preserves and transmits information about the frequency of co-occurrence of edges between nodes in good solutions - exactly the same information that the pheromone table encodes explicitly in the ACO algorithm. Figure 2 shows progress of one typical run for each population size and problem with Order (XP) and Uniform (NG) crossover. As can be seen the smaller populations make more rapid progress in the initial stages and all algorithms continue to discover improved solutions long into their runs.

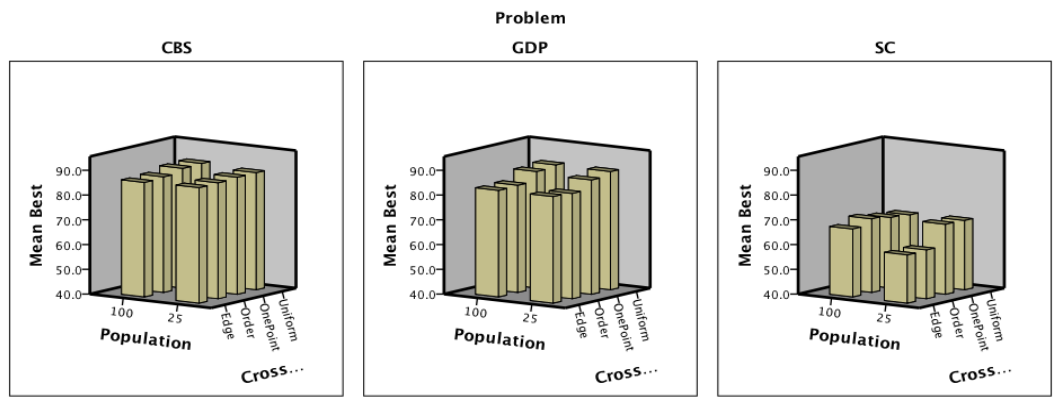

Figure 1: Mean best coupling $f_{CBO}$ achieved with EA using self-adaptive mutation, Px = 0.6. Edge and Order crossover are for XP representation, Uniform and One Point for NG

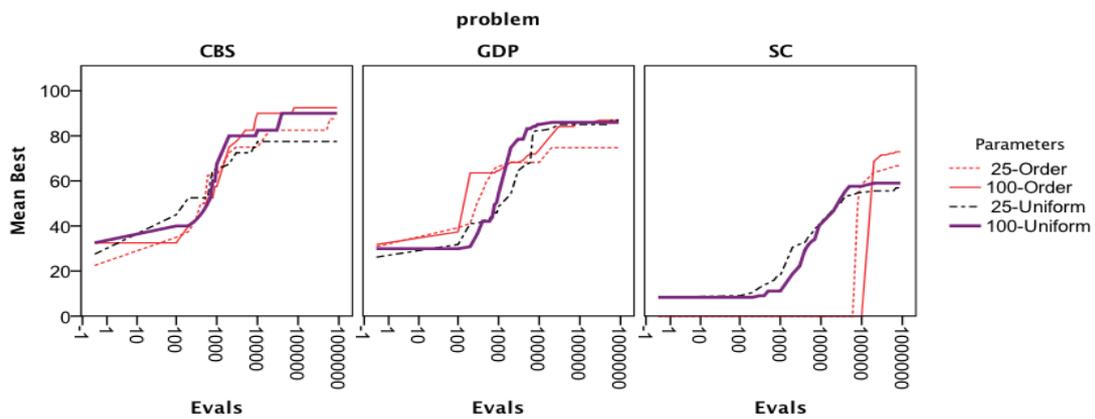

Figure 2: Progress of typical EA runs. Note logarithmic scale of x-axis

### 6.3 Ant Colony Optimization (ACO)

An illustration of sensitivity of the mean best $f_{CBO}$ values achieved for each design problem is shown in figure 3, for a value of ρ of 0.1 and a colony of 25 ants. To summarize the effects:

- α: performance increases as α increases from 0 to 1.0 – 1.5 but tails off thereafter;

- μ: performance increases as μ increases from zero to 3.0;

- ρ: little effect for CBS and GDP, but for SC performance increases as ρ increases from 0 to 1.0;

- numbers of ants appear to have little discernible affect for CBS and GDP, but some effect on SC.

Figure 3 shows mean values, but the complexity of the response surface was confirmed by a multiple linear regression analysis, where the goodness of fit was greatly improved by extending the model to include quadratic and two-way interactions between $\alpha$, $\mu$ and $\sigma$ for the three design problems.

In terms of the time taken to reach the best design solution, analysis shows that some degree of pheromone decay (i.e. $\rho > 0$) is necessary to achieve a plateau of fast performance (at higher values of $\alpha$ and $\mu$). This plateau effect is visible with a $\rho$ value of 0.01, but continues with increasing $\rho$ to 1.0. This suggests that a degree of pheromone decay is crucial in exploiting the search space by making the algorithm able to 'forget' design solutions of poor fitness, especially if they are infeasible (i.e. do not contain at least one attribute and one method).

Therefore, considering mean best coupling and number of evaluations together – the balance between exploration and exploitation - a value of $\alpha$ of 1.0 to 1.5 appears to be effective when combined with some degree of pheromone decay ($\rho >= 0.1$) and high values of pheromone update ($\mu = 3.0$) to balance the decay. Taking these values, figure 4 shows a typical ACO single run mean best $f_{CBO}$. It is observed that using 25 ants achieves mean best $f_{CBO}$ quicker than using 100 ants, although for CBS and GDP, using 100 ants achieves a superior $f_{CBO}$ after further evaluations.

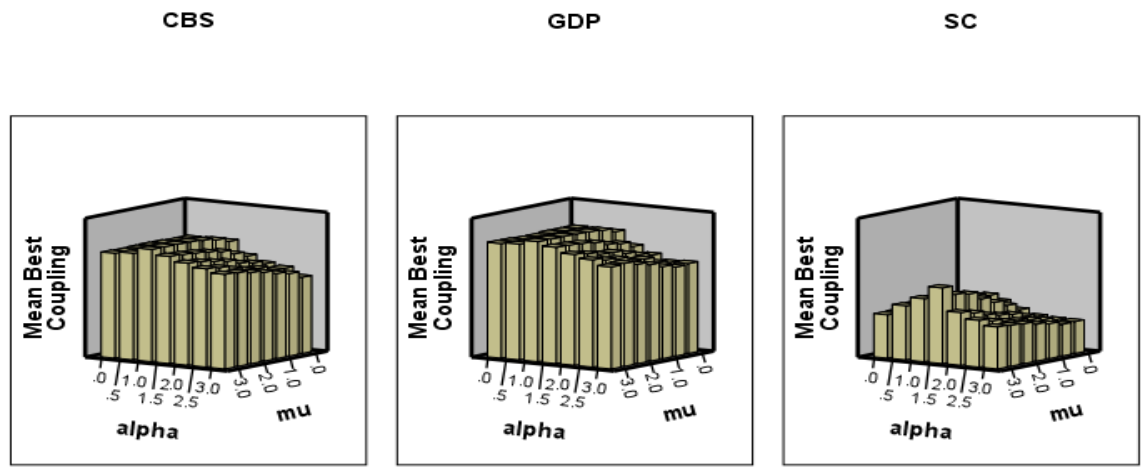

Figure 3: Sensitivity of $f_{CBO}$ values to parameters. 25 ants, $\rho = 0.1$

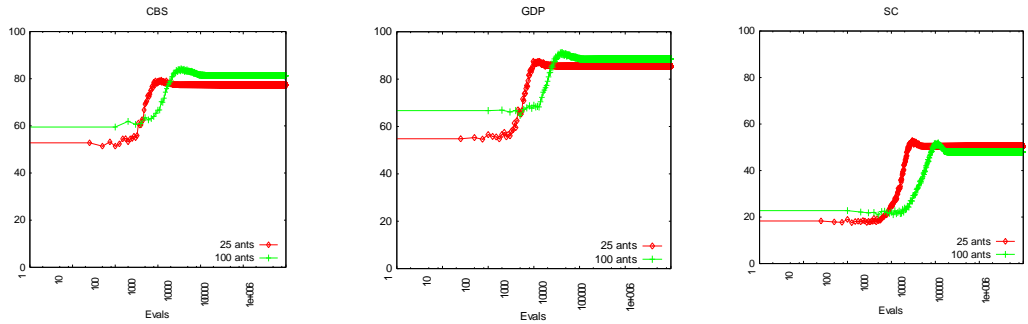

Figure 4: Progress of typical ACO runs. Note logarithmic scale of x-axis

Although ACO sensitivity to parameter values is undesirable from a robustness perspective, the "sweet spot' is the same for all three problems. With these settings, mean best $f_{CBO}$ values of 86.17, 93.39 and 47.20 for CBS, GDP and SC respectively compare favorably those of the manual design for CBS and GDP, although noticeably less so for SC, which prompts further analysis.

### 6.4 Comparative Analysis

Table 4 shows a comparison of the performance of the "best" version of the three search algorithms as identified above. The standard deviations are all in the range [0.5,5] except for GLS-XP (35-36) and GLS-NG on SC (27). Values in bold indicate the rankings per-problem according to the groups where the results are statistically significantly different. Given the budget of one million evaluations, it should be viewed as an indication of the ability of the algorithms to search landscapes induced by different representations for this problem. Since we have shown elsewhere that coupling is a correlated with human design judgment, if algorithms do not do well here, then there is little point subjecting humans to interacting with them. In most cases the metaheuristics create solutions with lower coupling than manual designs. Although GLS-NG does well on CBS and GDP, its low reliability on the SC appears to make it unsuitable for interactive applications. These results reveal an interesting pattern. Although the NG landscape appears to be far more amenable to search by GLS or the EA, the results for the ACO are better (although not significantly so for 50 runs) despite the fact that the ACO is searching the far more multi-modal and redundant working on the XP landscape. The exception to this is the SC problem. There are two possible reasons for this: the first is that the problem has more variables and that ACO does not scale. The second is that since it has more classes (16 as opposed to the 5 each for CBS and GSP) there are a far higher proportion of infeasible solutions, and the problem lies with the constraint handling.

Table 4: Comparison of Mean Best Coupling ($f_{CBO}$). Values in bold are statistically significant rankings.

|     | Manual | GLS (XP) | EA (XP) | ACO | GLS (NG) | EA (NG) |
|-----|--------|----------|---------|-----|----------|---------|
| CBS | 84.6   | 62.35, **5** | 82.10, **1=** | 90.00, **1=** | 87.25, **1=** | 88.80, **1=** |
| GDP | 70.3   | 57.61, **5** | 77.46, **4**  | 96.20, **1**  | 87.35, **1=** | 88.07, **1=** |
| SC  | 54.8   | 0.0, **5**   | 42.68, **4**  | 49.76, **3**  | 60.89, **2**  | 67.74, **1**  |

## 6.5 Constraint Handling

To investigate the worse behavior on the SC problem we ran ACO and EA-NG to produce designs with 5 classes rather than 16. In each case mean best $f_{CBO}$ values of over 90 were observed – showing that the difficulty is one of the proportion of the search space that is infeasible, rather than its size.

As described above, after an offspring solution (EA) or new solution path (ACO) is generated, a check is made to ensure that it contains at least one method and one attribute. In the indirect approach used so far, infeasible solutions are assigned a fitness 0.0. Although direct methods for constraint handling have been reported to be more effective (Eiben and Smith, 2003), they would normally require highly specialized operators, sophisticated repair mechanisms which would mitigate against the use of these meta-heuristics as general purposes engines for interactive search. We therefore implemented the simplest form of "direct" approach in the EA and ACO– each newly created solution is checked, and regenerated until a feasible one is created. In the ACO the number of repeated is capped at 50. Results with the ACO show that for the CBS and GDP problems the indirect approach leads to statistically significantly better results on the CBS, GDP and 5-class version of the SC problems, while for the 16-class SC the difference is not statistically significant. The differences between the numbers of evaluations to best solution are also not statistically significant. We hypothesize that direct method performs less well because enforcing validity in early generations increases the probability of creating redundant versions of effectively the same design, which will confuse the process whereby the pheromone table adapts to model the structure of the underlying problem.

In contrast to this, as shown in table 5, the beneficial results from implementing the direct mechanism in the EA are dramatic, especially as the scale of the problem, and hence the proportion of infeasible solutions increases, although notably, the less redundant NG representation still gives the best results.

Table 5: Comparison of Mean Best Coupling ($f_{CBO}$) with indirect and direct constraint handling. Values in bold are statistically significant rankings.

|     | Indirect-XP | Direct-XP | Indirect-NG | Direct-NG |
| --- | --- | --- | --- | --- |
| CBS | 82.10 (3.21), **4** | 87.95 (1.31), **1=** | 88.80 (2.49), **1=** | 88.55 (2.58), **1=** |
| GDP | 77.46 (2.75), **4** | 83.10 (2.48), **3** | 88.07 (3.69), **1=** | 88.36 (3.44), **1=** |
| SC  | 42.68 (5.79), **4** | 73.26 (2.09), **1=** | 67.73 (5.41), **3** | 71.92 (3.77), **1=** |

## 7. Interactive Search Simulation

The results in Section 6 concerned the ability of the meta-heuristics to reliably locate good solutions for the software design problem given a large amount of evaluations. While this is useful to rule out some methods (e.g. GLS), the overall ranking of algorithms is only really relevant if a surrogate fitness model is available, since in practice humans can only evaluate a few dozens of solutions before fatigue and lack of engagement sets in. Thus we next compare the performance in a more time-limited scenario, leveraging our previous work to calculate fitness as a stochastic weighted sum of coupling and two elegance metrics to simulate the effect of user evaluation. Since we would not ask users to evaluate infeasible solutions, and initial experimentation showed that the effect of redundancy was much less with the smaller population sizes appropriate to visual display, we used direct constraint handling, i.e. infeasible solutions are re-created until valid.

Table 6 shows the results of running the three meta-heuristics ACO, EA-NG and EA-XP with population size 10, run for a maximum of 250 evaluations. The results shown are for the best individual found according to the $f_{MO}$ metric averaged over 50 runs. The negative $f_{ATMR}$ values for ACO generating solutions to SC arise from the way that the maximization function was calculated using a scale factor of R = 6.0: the algorithm repeatedly identified solutions with low coupling and high NAC elegance, but high variability between classes of the attribute-methods ratio amongst classes in the solutions found. The superior performance listed for on almost every metric of the ACO algorithm, is confirmed by ANOVA followed by Tamhane's post-hoc test (since the variances are not equal).
While the coupling values obtained in a limited number of evaluations are inferior to the manual designs, the elegance values are comparable, and of course even 250 evaluations reflects less human effort than undertaking the manual design process. However, this does suggest that if it was intended to use $f_{MO}$ as a surrogate fitness measure, it would be worth investigating a non-linear function of $f_{CBO}$ to place more emphasis on minimizing coupling.

Table 6: Comparison of mean (std. dev.) results for best solutions found in simulation of interactive behaviour. Bold type indicates statistically significant best per row, i.e., problem-metric combination .

| Problem | Metric | EA-NG | EA-XP | ACO |
|---|---|---|---|---|
| CBS | $f_{MO}$ | 40.06 (5.79) | 45.73 (5.61) | **60.35 (3.88)** |
| | $f_{CBO}$ | 49.80 (7.28) | **56.55 (7.57)** | 48.26 (12.2) |
| | $f_{NAC}$ | 76.57 (6.32) | 46.82 (10.49) | **99.30 (0.08)** |
| | $f_{ATMR}$ | 92.21 (4.13) | 93.36 (3.89) | **97.34 (0.56)** |
| | AES | 210 (45) | 199 (54) | **140 (69)** |
| GDP | $f_{MO}$ | 31.73 (4.10) | 44.20 (6.94) | **57.16 (3.21)** |
| | $f_{CBO}$ | 39.50 (5.13) | 55.10 (8.68) | 59.12 (11.15) |
| | $f_{NAC}$ | 65.27 (11.57) | 4.46 (10.26) | **95.18 (0.77)** |
| | $f_{ATMR}$ | 70.82 (12.65) | 69.91 (13.26) | **94.64 (0.55)** |
| | AES | 230 (24) | 222 (31) | **127 (74)** |
| SC | $f_{MO}$ | 13.36 (2.03) | 15.04 (3.44) | **31.29 (3.41)** |
| | $f_{CBO}$ | 16.5 (2.55) | 18.55 (4.25) | **39.25 (11.93)** |
| | $f_{NAC}$ | 79.85 (2.98) | 73.328 (14.88) | **84.75 (2.01)** |
| | $f_{ATMR}$ | **78.92 (4.62)** | 76.57 (6.09) | -14.14 91.32 ) |
| | AES | 224 (38) | 210 (51) | **110 (73)** |

## 8. Analysis

### 8.1 Scalability

In contrast to Stochastic Local Search, both population-based search algorithms (EAs and ACO) afford a good degree of scalability with respect to the numbers of elements (attributes and methods) present in the interactive software design problem. For the CBS and GDP design problems, both EAs and ACO discover design solutions of superior fitness compared to the hand-crafted, manual design, for both single-objective search and multi-objective interactive search simulation. Indeed, with high numbers of elements, ACO outperforms the EAs. However, the ACO struggled to achieve superior fitness with the scale of the SC design problem. Further analysis revealed that the crucial factor affecting scalability is the proportion of infeasible solutions in the search space, which reflects the constraint that each class must have at least one method and attribute. In the results above we have examined SC with 16 classes, to permit comparison with the manual design. However, the constraint could be relaxed by allowing completely empty classes. This would provide a means of designing solutions with variable numbers of classes, so avoiding the need to pre-specify this important aspect of the design structure.

With respect to 'freezing' of partial design solutions, both population-based search algorithms offer the opportunity to permanently recording individual classes, or groups of classes. While the mechanisms for freezing (and unfreezing) parts of a design would require a major specialized adaptation of the EA, it is easily achieved in an ACO by simply setting the relevant pheromones to an arbitrary high value. Although not directly investigated in this paper due to simulation difficulties, this may nevertheless be a fruitful tactic in any future interactive studies.

## 8.2 Robustness

In line with previous findings (Stone and Smith 2002, Serpell and Smith 2010), results of EAs show that self-adapting mutation enables a good degree of robustness, making the algorithm insensitive to crossover probabilities or selection pressure. Conversely, it is evident that ACO is sensitive to parameter values, although once tuned, the same set of parameter values ($\alpha = 1.0\text{-}1.5$, $\mu = 3.0$, $\sigma = 0.1$) produces good results across all three software design problems. While the ACO graph representation appears more robust to varying numbers of elements in the software design solutions, the EA with integer-based naïve grouping representation appears more robust to varying numbers of classes.

## 8.3 'Off-the-Shelf' Availability

As described in section 4.3, key factors affecting the application of 'off-the-shelf' frameworks and toolkits for EAs and ACO appear to be (i) their ability to easily adapt the specifics of the search problem at hand, and (ii) available documentation and support. In the case of the software design problem, the specific constraint of the 'at least one attribute and one method' is not well catered for by the EA or ACO frameworks. In one sense this is not entirely surprising as software development frameworks are necessarily generic. In addition, available documentation, especially for ACO, is not sufficiently comprehensive. An additional although perhaps less important factor might be the choice of programming language provided by the framework. For example, ACO frameworks such as ACO (2012) focus on the C and C++ programming language which may or may not meet the portability and interoperability requirements of GUI-driven interactive meta-heuristic search engine.

## 8.4 Constraint Handling

Handling the constraint of each class containing at least one attribute and at least one method is a significant constraint affecting the performance of all the meta-heuristic search algorithms investigated. A number of tactics to deal with constraint present themselves. Firstly, the constraint can simply be ignored in search but invalid design solutions attract zero fitness. Secondly, the search algorithm can be adapted and specialized to prevent the construction of invalid design solutions. Thirdly, the search algorithm can also be adapted to repeat construction of a design solution until a valid solution appears. Much of the investigation in this paper centers on the use of the first tactic, which appears to be effective. Indeed, perhaps surprisingly, results of section 6.1.5 indicate that introducing constraint handling into the ACO algorithm produces inferior results from the point of view of an interactive design simulation. We hypothesize that this arises from the redundancy of the XP representation, and preliminary experimentation with decreasing population sizes for the MO search seemed to confirm our hypothesis.

## 9 Conclusions

This paper seeks to challenge the largely historical adoption of evolutionary computing as the basis of an engine for interactive search in early lifecycle software design. Greedy local search, evolutionary algorithms and ant colony optimization have been compared and the performance of the population-based algorithms has been found superior to single-parent stochastic local search. Indeed, experimental results show that population-based search produces software design solutions of fitness values superior to those of the hand-crafted design solutions.

Given a large computational budget – for example if a surrogate fitness model was available so only occasional user evaluation were required - the EA with the integer NG representation comes out as the clear favorite in terms of optimization performance. It is far more robust to parameter settings and the methods used for constraint handling.

However, if a wholly interactive search is required, thus small populations that can easily be visualized side-by-side, and a more limited computational budget, then a very different picture emerges.   In this case the use of an ACO, with each ant forced to recreate its solution path until valid, emerges as finding higher quality solutions, and in around half the time of the EAs. Moreover, the simple modifications that would be required to allow user-friendly modifications such as the ability to "freeze" certain classes, or coalesce others also point to the use of interactive ACOs for larger problems.